\documentclass{article}

\usepackage{microtype}
\usepackage{graphicx}
\usepackage{subfigure}
\usepackage{booktabs} 
\usepackage{amsmath}
\usepackage{amssymb}
\usepackage{amsthm}
\usepackage{bm}
\usepackage{bbm}

\usepackage{color}
\usepackage{adjustbox}
\usepackage{enumitem}
\usepackage{floatrow}
\usepackage{array}
\usepackage{multirow}
\usepackage{colortbl}
\usepackage{subfig}
\usepackage{xcolor}
\usepackage{caption}
\usepackage{soul}
\usepackage[textsize=scriptsize]{todonotes}

\definecolor{grey}{rgb}{0.8,0.8,0.8}
\definecolor{aqua}{rgb}{0, 1, 1}
\definecolor{steel}{rgb}{0.2734, 0.5078, 0.7031}
\definecolor{slate}{rgb}{0.1836, 0.3086, 0.3086}

\newcommand{\hlg}[2]{\setlength{\fboxsep}{0.3pt}\colorbox{green!#2}{\rule[-.05\baselineskip]{0pt}{.7\baselineskip}{#1}}}
\newcommand{\hlr}[2]{\setlength{\fboxsep}{0.3pt}\colorbox{red!#2}{\rule[-.05\baselineskip]{0pt}{.7\baselineskip}{#1}}}
\newcommand{\hlb}[2]{\setlength{\fboxsep}{0.3pt}\colorbox{aqua!#2}{\rule[-.05\baselineskip]{0pt}{.7\baselineskip}{#1}}}

\newcommand{\algname}{\textsc{InvRat }}
\newcommand{\algnamens}{\textsc{InvRat}}

\newtheorem{theorem}{Theorem}

\DeclareMathOperator*{\argmin}{\arg\!\min}
\DeclareMathOperator*{\argmax}{\arg\!\max}
\newcommand{\e}[1]{{\small $#1$}}

\usepackage{hyperref}

\usepackage[accepted]{icml2019}

\icmltitlerunning{Invariant Rationalization}

\begin{document}

\twocolumn[
\icmltitle{Invariant Rationalization}



\icmlsetsymbol{equal}{*}

\begin{icmlauthorlist}
\icmlauthor{Shiyu Chang$^1$}{equal}
\icmlauthor{Yang Zhang$^1$}{equal}
\icmlauthor{Mo Yu$^2$}{equal}
\icmlauthor{Tommi S. Jaakkola$^3$}{} \\
\vspace*{0.05in}
\icmlauthor{\textnormal{$^1$MIT-IBM Watson AI Lab \qquad $^2$IBM Research \qquad $^3$CSAIL MIT}}{}\\
\vspace*{0.05in}
\icmlauthor{\textnormal{\small{\texttt{\{shiyu.chang, yang.zhang2\}@ibm.com}\quad \texttt{yum@us.ibm.com} \quad \texttt{tommi@csail.mit.edu}}}}{}
\end{icmlauthorlist}




\vskip 0.3in
]



\printAffiliationsAndNotice{\icmlEqualContribution} 

\begin{abstract}

Selective rationalization improves neural network interpretability by identifying a small subset of input features — the rationale — that best explains or supports the prediction. A typical rationalization criterion, \emph{i.e.} maximum mutual information (MMI), finds the rationale that maximizes the prediction performance based only on the rationale. However, MMI can be problematic because it picks up spurious correlations between the input features and the output.  Instead, we introduce a game-theoretic invariant rationalization criterion where the rationales are constrained to enable the same predictor to be optimal across different environments. 
We show both theoretically and empirically that the proposed rationales can rule out spurious correlations, generalize better to different test scenarios, and align better with human judgments.
Our data and code are available.\footnote{\scriptsize{\url{https://github.com/code-terminator/invariant_rationalization}}.}

\end{abstract}

\section{Introduction}
\label{sec:intro}
A number of selective rationalization techniques \cite{lei2016rationalizing,li2016understanding,chen2018learning,chen2018shapley, yu2018learning,  yu2019rethinking, chang2019game} have been proposed to explain the predictions of complex neural models. The key idea driving these methods is to find a small subset of the input features --  \emph{rationale} -- that suffices on its own to yield the same outcome. In practice, rationales that remove much of the spurious content from the input, \emph{e.g.}, text, could be used and examined as justifications for model's predictions.

The most commonly-used criterion for rationales is the maximum mutual information (MMI) criterion. In the context of NLP, it defines rationale as the subset of input text that maximizes the mutual information between the subset and the model output, subject to the constraint that the selected subset remains within a prescribed length.  Specifically, if we denote the random variables corresponding to input as \e{\bm X}, rationales as \e{\bm Z} and the model output as \e{Y}, then the MMI criterion finds the explanation \e{\bm Z =\bm Z(\bm X)} that yields the highest prediction accuracy of \e{Y}.

MMI criterion can nevertheless lead to undesirable results in practice. It is prone to highlighting spurious correlations between the input features and the output as valid explanations. While such correlations represent statistical relations present in the training data, and thus incorporated into the neural model, the impact of such features on the true outcome (as opposed to model's predictions) can change at test time. In other words, MMI may select features that do not explain the underlying relationship between the inputs and outputs even though they may still be faithfully reporting the model's behavior. We seek to modify the rationalization criterion to better tailor it to find causal features. 

\floatsetup[table]{capposition=bottom}
\begin{table*}[t]
	\small
	\begin{tabular}{p{\linewidth}}
        \emph{Beer - Smell}\hspace*{0pt}\hfill Label - Positive\\
		\arrayrulecolor{grey}  
		\midrule
        \hlr{375ml}{0} \hlr{corked}{0} \hlr{and}{0} \hlr{caged}{0} \hlr{bottle}{0} \hlr{with}{0} \hlr{bottled}{0} \hlr{on}{0} \hlr{date}{0} \hlr{november}{0} \hlr{30}{0} \hlr{2005}{0} \hlr{,}{0} \hlr{poured}{0} \hlr{into}{0} \hlr{snifter}{0} \hlr{at}{0} \hlr{brouwer}{0} \hlr{'s}{0} \hlr{,}{0} \hlr{reviewed}{0} \hlr{on}{0} \hlr{5/15/11}{0} \hlr{.}{0} \hlr{aroma}{0} \hlr{:}{0} \hlr{pours}{0} \hlr{a}{0} \hlr{clear}{0} \hlr{golden}{0} \hlr{color}{0} \hlr{with}{0} \hlr{orange}{0} \hlr{hues}{0} \hlr{and}{0} \hlr{a}{0} \hlr{whitish}{0} \hlr{head}{0} \hlr{that}{0} \hlr{leaves}{0} \hlr{some}{0} \hlr{lacing}{0} \hlr{around}{0} \hlr{glass}{0} \hlr{.}{0} \hlr{smell}{0} \hlr{:}{0} \hlg{\ul{\textbf{lots}}}{20} \hlg{\ul{\textbf{of}}}{20} \hlg{\ul{\textbf{barnyaardy}}}{20} \hlg{\ul{\textbf{funk}}}{20} \hlg{\ul{\textbf{with}}}{20} \hlg{\ul{\textbf{tons}}}{20} \hlg{\ul{\textbf{of}}}{20} \hlg{\ul{\textbf{earthy}}}{20} \hlg{\ul{\textbf{aromas}}}{20} \hlg{\ul{\textbf{,}}}{20} \hlg{\ul{\textbf{grass}}}{20} \hlg{\ul{\textbf{and}}}{20} \hlg{\ul{\textbf{some}}}{20} \hlg{\ul{\textbf{lemon}}}{20} \hlg{\ul{\textbf{peel}}}{20} \hlr{.}{0} \hlr{palate}{0} \hlr{:}{0} \hlr{\ul{\textbf{similar}}}{20} \hlr{\ul{\textbf{to}}}{20} \hlr{\ul{\textbf{the}}}{20} \hlr{\ul{\textbf{aroma}}}{20} \hlr{\ul{\textbf{,}}}{20} \hlr{\ul{\textbf{lots}}}{20} \hlr{\ul{\textbf{of}}}{20} \hlr{\ul{\textbf{funk}}}{20} \hlr{\ul{\textbf{,}}}{20} \hlr{\ul{\textbf{lactic}}}{20} \hlr{\ul{\textbf{sourness}}}{20} \hlr{\ul{\textbf{,}}}{20} \hlr{\ul{\textbf{really}}}{20} \hlr{\ul{\textbf{earthy}}}{20} \hlr{\ul{\textbf{with}}}{20} \hlr{\ul{\textbf{citrus}}}{20} \hlr{\ul{\textbf{notes}}}{20} \hlr{\ul{\textbf{and}}}{20} \hlr{\ul{\textbf{oak}}}{20} \hlr{\ul{\textbf{.}}}{20} \hlr{\ul{\textbf{many}}}{20} \hlr{\ul{\textbf{layers}}}{20} \hlr{\ul{\textbf{of}}}{20} \hlr{\ul{\textbf{intriguing}}}{20} \hlr{\ul{\textbf{earthy}}}{20} \hlr{\ul{\textbf{complexities}}}{20} \hlr{.}{0}
        \hlr{overall}{0} \hlr{:}{0} \hlb{\ul{\textbf{very}}}{20} \hlb{\ul{\textbf{funky}}}{20} \hlb{\ul{\textbf{and}}}{20} \hlb{\ul{\textbf{earthy}}}{20} \hlb{\ul{\textbf{gueuze}}}{20} \hlb{\ul{\textbf{,}}}{20} \hlb{\ul{\textbf{nice}}}{20} \hlb{\ul{\textbf{and}}}{20} \hlb{\ul{\textbf{crisp}}}{20} \hlb{\ul{\textbf{with}}}{20} \hlb{\ul{\textbf{good}}}{20} \hlb{\ul{\textbf{drinkability}}}{20} \hlr{.}{0} \\
	\end{tabular}
	\vspace*{-0.15in}
    \captionof{figure}{\small{An example beer review and possible rationales explaining why the score on the smell aspect is positive. \hlg{\ul{\textbf{Green highlights}}}{20} the review on the smell aspect, which is the true explanation.  \hlr{\ul{\textbf{Red highlights}}}{20} the review on the taste aspect, which has a high correlation with the smell. \hlb{\ul{\textbf{Blue highlights}}}{20} the overall review, which summarizes all the aspects, including smell. All three sentences have high predictive powers of the smell score, but only the green sentence is the desired explanation.}}
    \label{fig:intro_example}
\end{table*}
\floatsetup[table]{capposition=top}

As an example, consider figure~\ref{fig:intro_example} that shows a beverage review which covers four aspects of beer: \emph{appearance}, \emph{smell}, \emph{palate}, and \emph{overall}. The reviewers also assigned a score to each of these aspects. Suppose we want to find an explanation supporting a positive score to \emph{smell}. The correct explanation should be the portion of the review that actually discusses smell, as highlighted in green. However, reviews for other aspects such as \emph{palate} (highlighted in red) may co-vary with \emph{smell} score since, as senses, smell and palate are related. The overall statement as highlighted in blue would typically also clearly correlate with any individual aspect score, including \emph{smell}. Taken together, sentences highlighted in green, red and blue would all be highly correlated with the positive score for \emph{smell}. As a result, MMI may select any one of them (or some combination) as the rationale, depending on precise statistics in the training data. Only the green sentence constitutes an adequate explanation. 

Our goal is to design a rationalization criterion that approximates finding causal features. While assessing causality is challenging, we can approximate the task by searching features that are \emph{invariant}. This notion was recently introduced in the context of invariant risk minimization (IRM) \cite{arjovsky2019invariant}. The main idea is to highlight spurious (non-causal) variation by dividing the data into different environments. The same predictor, if based on causal features, should remain optimal in each environment separately.

In this paper, we propose invariant rationalization (\algnamens), a novel rationalization scheme that incorporates the invariance constraint. We extend the IRM principle to neural predictions by resorting to a game-theoretic framework to impose invariance. Specifically, the proposed framework consists of three modules: a rationale generator, an environment-agnostic predictor as well as an environment-aware predictor. The rationale generator generates rationales \e{\bm Z} from the input \e{\bm X}, and both predictors try to predict \e{Y} from \e{\bm Z}. The only difference between the two predictors is that the environment-aware predictor also has access to which environment each training data point is drawn from. The goal of the rationale generator is to restrict the rationales in a manner that closes the performance gap between the two predictors while still maximizing the prediction accuracy of the environment-agnostic predictor.

We show theoretically that \algname can solve the invariant rationalization problem, and that the invariant rationales generalize well to unknown test environments in a well-defined minimax sense. We evaluate \algname on multiple datasets with false correlations. The results show that \algname does significantly better in removing false correlations and finding explanations that better align with human judgments. 

\section{Preliminaries: MMI and Its Limitation}
In this section, we will formally review the MMI criterion and analyze its limitation using a probabilistic model. Throughout the paper, upper-cased letters, \e{X} and \e{\bm X}, denote random scalars and vectors respectively; lower-cased letters, \e{x} and \e{\bm x}, denote deterministic scalars and vectors respectively; \e{H(\bm X)} denotes the Shannon entropy of \e{\bm X}; \e{H(Y | \bm X)} denotes the entropy of \e{Y} conditional on \e{\bm{X}}; \e{I(Y; \bm X)} denotes the mutual information. Without causing ambiguities, we use \e{p_{\bm X}(\cdot)} and \e{p(\bm X)} interchangeably to denote the probabilistic mass function of \e{\bm X}.

\subsection{Maximum Mutual Information Criterion}

The MMI objective can be formulated as follows. Given the input-output pairs \e{(\bm X, Y)}, MMI aims to find a rationale \e{\bm Z}, which is a masked version of \e{\bm X}, such that it maximizes the mutual information between \e{\bm Z} and \e{Y}. Formally,
\begin{equation}
\small
\max_{\bm m \in \mathcal{S}} I(Y; \bm Z) \quad \mbox{s.t. } \bm Z= \bm m \odot \bm X, \\
\label{eq:tao}
\end{equation}
where \e{\bm m} is a binary mask and \e{\mathcal{S}} denotes a subset of \e{\{0, 1\}^N} with a sparsity and a continuity constraints. \e{N} is the total length in \e{\bm X}.  We leave the exact mathematical form of the constraint set \e{\mathcal{S}} abstract here, and it will be formally introduced in section \ref{ssec:constraints}. \e{\odot} denotes the element-wise multiplication of two vectors or matrices. Since the mutual information measures the predictive power of \e{\bm Z} on \e{Y}, MMI essentially tries to find a subset of input features that can best predict the output \e{Y}.

\begin{figure}[t]
\centering
\includegraphics[width=0.65\linewidth]{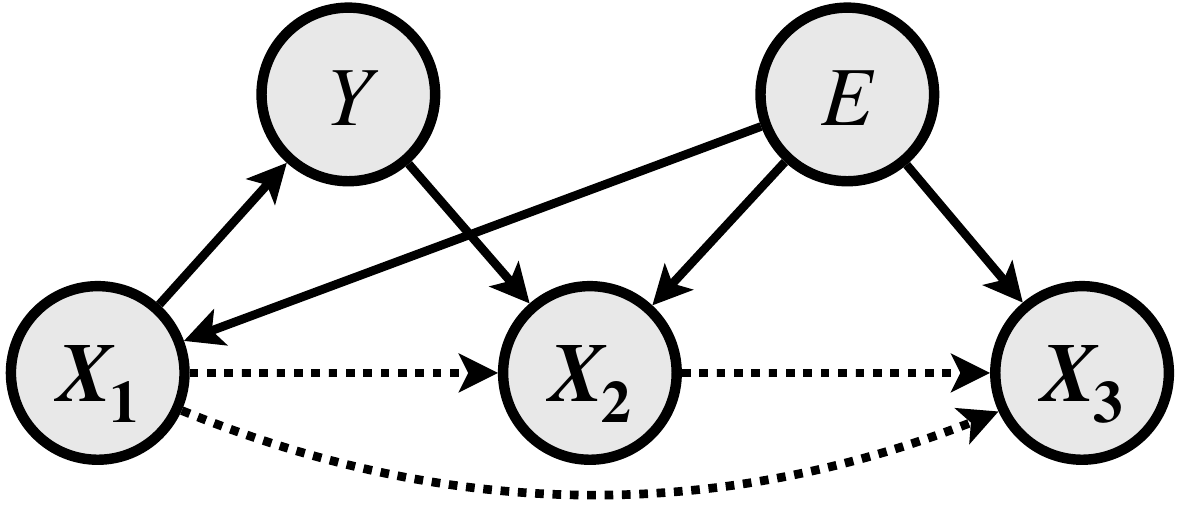}
\vspace*{-0.1in}
\caption{\small{A probabilistic model illustrating different parts of an input that have different probabilistic relationships with the model output \e{Y}.  A sentence \e{\bm X} can be divided into three variables \e{\bm X_1}, \e{\bm X_2} and \e{\bm X_3}.  All \e{\bm X_1}, \e{\bm X_2} and \e{\bm X_3} can be highly correlated with \e{Y}, but only \e{\bm X_1} is regarded as a plausible explanation.}}
\label{fig:model}
\end{figure}

\subsection{MMI Limitations}
\label{subsec:mmi_limit}

The biggest problem of MMI is that it is prone to picking up spurious probabilistic correlations, rather than finding the causal explanation. To demonstrate why this is the case, consider a probabilistic graph in figure~\ref{fig:model}, where \e{\bm X} is divided into three variables, \e{\bm X_1}, \e{\bm X_2} and \e{\bm X_3}, which represents the three typical relationship with \e{Y}: \e{\bm X_1} influences \e{Y}; \e{\bm X_2} is influenced by \e{Y}; \e{\bm X_3} has no direction connections with \e{Y}. The dashed arrows represent some additional probabilistic dependencies among \e{\bm X}. 
For now, we ignore \e{E}.

As observed from the graph, \e{\bm X_1} serves as the valid explanation of \e{Y}, because it is the true cause of \e{Y}. Neither \e{\bm X_2} nor \e{\bm X_3} are valid explanations. However, \e{\bm X_1}, \e{\bm X_2} and \e{\bm X_3} can all be highly predicative of \e{Y}, so the MMI criterion may select any of the three features as the rationale. Concretely, consider the following toy example with all binary variables. Assume \e{p_{\bm X_1}(1) = 0.5}, 
and
\begin{equation}
    \small
    p_{Y | \bm X_1}(1 | 1) = p_{Y | \bm X_1}(0 | 0) = 0.9, 
    \label{eq:x1_pred}
\end{equation}
which makes \e{\bm X_1} a good predictor of \e{Y}. Next, define the conditional prior of \e{\bm X_2} as
\begin{equation*}
    \small
    \begin{aligned}
    p_{\bm X_2 | Y}(1 | 1) = p_{\bm X_2 | Y}(0 | 0) = 0.9.
    \end{aligned}
\end{equation*}
According to the Bayes rule,
\begin{equation}
    \small
    p_{Y | \bm X_2}(1 | 1) = p_{Y | \bm X_2}(0 | 0) = 0.9, 
    \label{eq:x2_pred}
\end{equation}
which makes \e{\bm X_2} also a good predictor of \e{Y}. Finally, assume the conditional prior of \e{\bm X_3} is
\begin{equation*}
\small
\begin{aligned}
    &p_{\bm X_3 | \bm X_1, \bm X_2}(1 | 1, 1) = p_{X_3 | \bm X_1, \bm X_2}(0 | 0, 0) = 1, \text{ and } \\
    &p_{\bm X_3 | \bm X_1, \bm X_2}(1 | 0, 1) = p_{X_3 | \bm X_1, \bm X_2}(1 | 1, 0) = 0.5.
\end{aligned}
\end{equation*}
It can be computed that
\begin{equation}
    \small
    p_{\bm X_3 | Y}(1 | 1) = p_{\bm X_3 | Y}(0 | 0) = 0.9.
    \label{eq:x3_pred}
\end{equation}
In short, according to equations~\eqref{eq:x1_pred}, \eqref{eq:x2_pred} and \eqref{eq:x3_pred}, we have constructed a set of priors such that the predictive power of \e{\bm X_1}, \e{\bm X_2} and \e{\bm X_3} is \emph{exactly the same}. As a result, there is no reason for MMI to favor \e{\bm X_1} over the others. 

In fact, \e{\bm X_1}, \e{\bm X_2} and \e{\bm X_3} correspond to the three highlighted sentences in figure~\ref{fig:intro_example}. \e{\bm X_1} corresponds to the smell review (green sentence), because it represents the true explanation that influences the output decision. \e{\bm X_2} corresponds to the overall review (blue sentence), because the overall summary of the beer inversely influenced by the smell score.  Finally, \e{\bm X_3} corresponds to the palate review (red sentence), because the palate review does not have a direct relationship with the smell score. However, \e{\bm X_3} may still be highly predicative of \e{Y} because it can be strongly correlated with \e{\bm X_1} and \e{\bm X_2}. Therefore, we need to explore a novel rationalization scheme that can distinguish \e{\bm X_1} from the rest.

\section{Adversarial Invariant Rationalization}
\label{subsec:air}
In this section, we propose invariant rationalization, a rationalization criterion that can exclude rationales with spurious correlations, utilizing the extra information provided by an environment variable.  We will introduce \algnamens, a game-theoretic approach to solving the invariant rationalization problem. We will then theoretically analyze the convergence property and the generalizability of invariant rationales.

\subsection{Invariant Rationalization}

Without further information, distinguishing \e{\bm X_1} from \e{\bm X_2} and \e{\bm X_3} is a challenging task. However, this challenge can be resolved if we also have access to an extra piece of information: the environment. As shown in figure~\ref{fig:model}, an environment is defined as an instance of the variable \e{E} that impacts the prior distribution of \e{\bm X} \cite{arjovsky2019invariant}.  On the other hand, we make the same assumption as in IRM that the \e{p(Y | \bm X_1)} remains the same across the environments (hence there is no edge pointing from \e{E} to \e{Y} in figure~\ref{fig:model}), because \e{\bm X_1} is the true cause of \e{Y}. As we will show soon, \e{p(Y | \bm X_2)} and \e{p(Y | \bm X_3)} will \emph{not} remain the same across the environments, which distinguishes \e{\bm X_1} from \e{\bm X_2} and \e{\bm X_3}.

Back to the binary toy example in section~\ref{subsec:mmi_limit}, suppose there are two environments, \e{e_1} and \e{e_2}. In environment \e{e_1}, all the prior distributions are exactly the same as in section~\ref{subsec:mmi_limit}. In environment \e{e_2}, the priors are almost the same, except for the prior of \e{\bm X_1}. For notation ease, define \e{q_{\bm X}(\cdot)} as the probabilities under environment \e{e_2}, \emph{i.e.} \e{p_{\bm X| E}(\cdot | e_2)}. Then, we assume that
\begin{equation*}
    \small
    \begin{aligned}
    q_{\bm X_1}(1) = 0.6.
    \end{aligned}
\end{equation*}
It turns out that such a small difference suffices to expose \e{\bm X_2} and \e{\bm X_3}.  In this environment, \e{q(Y | \bm X_1)} is the same as in equation \eqref{eq:x1_pred} as assumed.  However, it can be computed that
\begin{equation*}
\small
    \begin{aligned}
    q_{Y | \bm X_2}(1 | 1) \approx 0.926, \quad q_{Y | \bm X_2}(0 | 0) \approx 0.867,\\
    q_{Y | \bm X_3}(1 | 1) \approx 0.912, \quad q_{Y | \bm X_3}(0 | 0) \approx 0.883, 
    \end{aligned}
\end{equation*}
which are different from equations \eqref{eq:x2_pred} and \eqref{eq:x3_pred}. Notice that we have not yet assumed any changes in the priors of \e{\bm X_2} and \e{\bm X_3}, which will introduce further differences. The fundamental cause of such differences is that \e{Y} is independent of \e{E} \emph{only when} conditioned on \e{\bm X_1}, so \e{p_{Y | \bm X_1}(\cdot | \cdot)} would not change with \e{E}. We call this property \emph{invariance}. However, the conditional independence does not hold for \e{\bm X_2} and \e{\bm X_3}.

Therefore, given that we have access to multiple environments during training, \emph{i.e.} multiple instances of \e{E}, we propose the invariant rationalization objective as follows:
\begin{equation}
\small
\max_{\bm m \in \mathcal{S}} I(Y; \bm Z) \quad \mbox{s.t. } \bm Z= \bm m \odot \bm X, \quad Y \perp E ~|~ \bm Z,
\label{eq:ir}
\end{equation}
where \e{\perp} denotes probabilistic independence. The only difference between equations \eqref{eq:tao} and \eqref{eq:ir} is that the latter has the invariance constraint, which is used to screen out \e{\bm X_2} and \e{\bm X_3}.  In practice, finding an eligible environment is feasible. In the beer review example in figure~\ref{fig:intro_example}, a possible choice of environment is the brand of beer, because different beer brands have different prior distributions of the review in each aspect -- some brands are better at the appearance, others better at the palate. Such variations in priors suffice to expose the non-invariance of the palate review or the overall review in terms of predicting the smell score.

\subsection{The \algname Framework}

\begin{figure}[t]
\centering
\includegraphics[width=\linewidth]{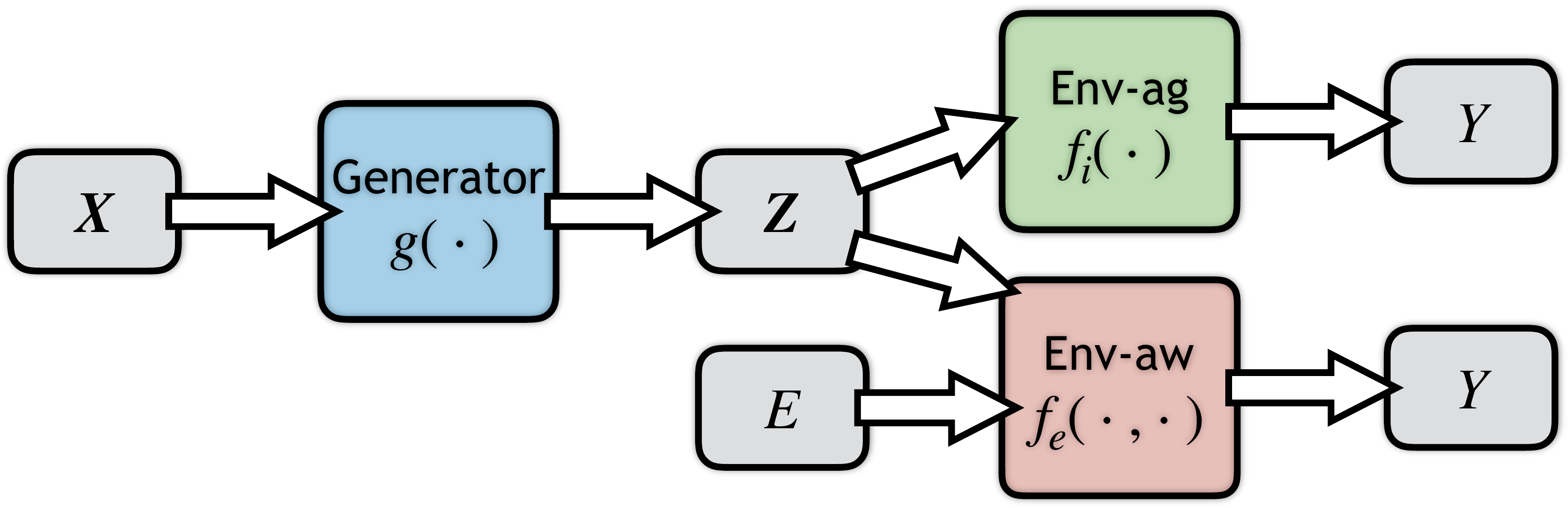}
\vspace*{-0.15in}
\caption{\small{The \algname framework with three players: the rationale generator, environment-agnostic and -aware predictors. }}
\label{fig:method}
\end{figure}

The constrained optimization in equation \eqref{eq:ir} is hard to solve in its original form. \algname introduces a game-theoretic framework, which can approximately solve this problem. Notice that the invariance constraint can be converted to a constraint on entropy, \emph{i.e.},
\begin{equation}
    \small
    \begin{aligned}
    Y \perp E ~|~ \bm Z \Leftrightarrow H(Y|\bm Z, E) = H(Y | \bm Z), 
    \end{aligned}
    \label{eq:invariance}
\end{equation}
which means if \e{\bm Z} is invariant, \e{E} cannot provide extra information beyond \e{\bm Z} to predict \e{Y}. Guided by this perspective, \algname consists of three players, as shown in figure \ref{fig:method}:

\begin{itemize}
    \setlength\itemsep{-0.2em}
    \item an environment-agnostic/-independent predictor \e{f_i(\bm Z)};
    \item an environment-aware predictor \e{f_e(\bm Z, E)}; and
    \item a rationale generator, \e{g(\bm X)}.
\end{itemize}

The goal of the environment-agnostic and environment-aware predictors is to predict \e{Y} from the rationale \e{\bm Z}. The only difference between them is that the latter has access to \e{E} as another input feature but the former does not. Formally, denote \e{\mathcal{L}(Y; f)} as the cross-entropy loss on a single instance. Then the learning objective of these two predictors can be written as follows.
\begin{equation}
    \small
    \mathcal{L}^*_i = \min_{f_i(\cdot)} \mathbb{E}[\mathcal{L}(Y; f_i(\bm Z))], ~ \mathcal{L}^*_e = \min_{f_e(\cdot, \cdot)} \mathbb{E}[\mathcal{L}(Y; f_e(\bm Z, E))],
    \label{eq:obj_pred}
\end{equation}
where \e{\bm Z = g(\bm X)}.  The rationale generator generates \e{\bm Z} by masking \e{\bm X}. The goal of the rationale generator is also to minimize the invariance prediction loss \e{\mathcal{L}^*_i }. However, there is an additional goal to make the gap between \e{\mathcal{L}^*_i} and \e{\mathcal{L}^*_e} small. Formally, the objective of the generator is as follows:
\begin{equation}
    \small
    \min_{g(\cdot)} \mathcal{L}^*_i + \lambda h(\mathcal{L}^*_i - \mathcal{L}^*_e), 
    \label{eq:obj_gen}
\end{equation}
where \e{h(t)} a convex function that is monotonically increasing in \e{t} when \e{t<0}, and \emph{strictly} monotonically increasing in \e{t} when \e{t\geq 0}, \emph{e.g.}, \e{h(t) = t} and \e{h(t)=\textrm{ReLU}(t)}.

\subsection{Convergence Properties}
This section justifies that equations \eqref{eq:obj_pred} and \eqref{eq:obj_gen} can solve equation~\eqref{eq:ir} in its Lagrangian form.  If the representation power of \e{f_i(\cdot)} and \e{f_e(\cdot, \cdot)} is sufficient, the cross-entropy loss can achieve its entropy lower bound, \emph{i.e.},
\begin{equation*}
    \small
    \begin{aligned}
    \mathcal{L}_i^* = H(Y | \bm Z), \quad \mathcal{L}_e^* = H(Y | \bm Z, E).
    \end{aligned}
\end{equation*}
Notice that the environment-aware loss should be no greater than the environment-agnostic loss, because of the availability of more information, \emph{i.e.}, \e{H(Y | \bm Z) \geq H(Y | \bm Z, E)}. Therefore, the invariance constraint in equation~\eqref{eq:invariance} can be rewritten as an inequality constraint:
\begin{equation}
\small
    H(Y | \bm Z) = H(Y | \bm Z, E) \Leftrightarrow H(Y | \bm Z) \leq H(Y | \bm Z, E).
\end{equation}
Finally, notice that \e{I(Y; \bm Z) = H(Y) - H(Y | \bm Z)}. Thus the objective in equation~\eqref{eq:obj_gen} can be regarded as the Lagrange form of equation~\eqref{eq:ir}, with the constraint rewritten as an inequality constraint
\begin{equation}
\small
    h(H(Y | \bm Z) - H(Y | \bm Z, E)) \leq h(0).
    \label{eq:ineq_contraint}
\end{equation}
According to the KKT conditions, \e{\lambda > 0} when equation \eqref{eq:ineq_contraint} is binding.  Moreover, the objectives in equations~\eqref{eq:obj_pred} and \eqref{eq:obj_gen} can be rewritten as a minimax game
\begin{equation}
    \small
    \min_{g(\cdot), f_i(\cdot)} \max_{f_e(\cdot, \cdot)} \mathcal{L}_i(g, f_i) + \lambda h(\mathcal{L}_i(g, f_i) - \mathcal{L}_e(g, f_e)),
    \label{eq:minimax}
\end{equation}
where
\begin{equation*}
    \small
    \begin{aligned}
    \mathcal{L}_i(g, f_i) = \mathbb{E}[\mathcal{L}(Y; f_i(\bm Z))], ~~ \mathcal{L}_e(g, f_e) = \mathbb{E}[\mathcal{L}(Y; f_e(\bm Z, E))].
    \end{aligned}
\end{equation*}
Therefore, the generator plays a co-operative game with the environment-agnostic predictor, and an adversarial game with the environment-aware predictor. The optimization can be performed using alternate gradient descent/ascent.

\subsection{Invariance and Generalizability}

In our previous discussions, we have justified the invariant rationales in the sense that it can uncover consistent and causal explanations and leave out spurious statistical correlations. In this section, we further justify invariant rationale in terms of generalizability. We consider two sets of environments, a set of training environments \e{\{e_t\}} and a test environment \e{e_a}. Only the training environments are accessible during training. The prior distributions in the test environment are completely unknown. The question we want to ask is: does keeping the invariant rationales and dropping the non-invariant rationales improve the generalizability in the unknown test environment?

Assume that 1) the training data are sufficient, 2) the predictor is environment-agnostic, 3) the predictor has sufficient representation power, and 4) the training converges to the global optimum. Under these assumptions, any predictor is able to replicate the training set distribution (with all the training environments mixed) \e{p(Y | \bm Z, E\in\{e_t\})}, which is optimal under the cross-entropy training objective. In the test environment \e{e_a}, the cross-entropy loss of this predictor is given by 
\begin{equation*}
\small
    \begin{aligned}
    \mathcal{L}^*_\textrm{test}(\bm Z) = H(p(Y | \bm Z, e_a) ; p(Y | \bm Z, \{e_t\})).
    \end{aligned}
\end{equation*}
where \e{p(Y | \bm Z, \{e_t\})} is short for \e{p(Y | \bm Z, E\in \{e_t\})}.  \e{\mathcal{L}^*_\textrm{test}(\bm Z)} cannot be evaluated because the prior distribution in the test environment is unknown. Instead, we consider the worst scenario. For notational ease, we introduce the following shorthand for the test environment distributions:
\begin{equation*}
\small
    \begin{aligned}
    &\pi_1(\bm x_1) = p_{\bm X_1 | E}(\bm x_1 | e_a),\\
    &\pi_2(\bm x_2 | \bm x_1, y) = p_{\bm X_2 | \bm X_1, Y, E}(\bm x_2 | \bm x_1, y, e_a),\\
    &\pi_3(\bm x_3 | \bm x_1, \bm x_2) = p_{\bm X_3 | \bm X_1, \bm X_2, E}(\bm x_3 | \bm x_1, \bm x_2, \cdot, e_a).
    \end{aligned}
\end{equation*}
For the selected rationale \e{\bm Z}, we consider an adversarial test environment (hence the notation \e{e_a}), which chooses \e{\pi_1}, \e{\pi_2} and \e{\pi_3} to maximize \e{\mathcal{L}^*_\textrm{test}(\bm Z; \pi_1, \pi_2, \pi_3)} (note that \e{\mathcal{L}^*_{\textrm{test}}(\bm Z)} is a function of \e{\pi_1}, \e{\pi_2}, and \e{\pi_3}). 
The following theorem shows that the minimizer of this adversarial loss is the invariant rationale \e{\bm X_1}.

\begin{theorem}
Assume the probabilistic graph in figure~\ref{fig:model} and that there are two environments \e{e_t} and \e{e_a}. \e{\bm Z = \bm X_1} achieves the saddle point of the following minimax problem
\begin{equation*}
    \small
    \begin{aligned}
    \min_{\bm Z \in \mathcal{X}} \max_{\pi_1, \pi_2, \pi_3} \mathcal{L}^*_\textrm{test}(\bm Z;\pi_1, \pi_2, \pi_3),
    \end{aligned}
\end{equation*}
where \e{\mathcal{X}} denotes the power set of \e{[\bm X_1, \bm X_2, \bm X_3]}.
\label{thm:invariance}
\end{theorem}
The proof is provided in the appendix \ref{appendix:proof}. Theorem \ref{thm:invariance} shows the nice property of the invariance rationale that it minimizes the risk under the most adverse test environment. 

\subsection{Incorporating Sparsity and Continuity Constraints}
\label{ssec:constraints}

The sparsity and continuity constraint \e{\bm m \in \mathcal{S}} (equation~\eqref{eq:ir}) stipulates that the total number of \e{1}'s in \e{\bm m} should be upper bounded and contiguous. There are two ways to implement the constraints.

\textbf{Soft constraints: }Following \citet{chang2019game}, we can add another two Lagrange terms to equations \eqref{eq:minimax}:
\begin{equation}
\small
    \mu_1 \bigg\lvert\frac{1}{N}\mathbb{E}[ \lVert \bm m \rVert_1] -\alpha \bigg\rvert + \mu_2 \mathbb{E} \bigg[\sum_{n=2}^N \lvert \bm m_n - \bm m_{n-1} \rvert \bigg],
    \label{eq:soft_constraint}
\end{equation}
where \e{\bm m_n} denotes the \e{n}-th element of \e{\bm m}; \e{\alpha} is a predefined sparsity level. \e{\bm m} is produced by an independent selection process \cite{lei2016rationalizing}. This method is flexible, but requires sophisticated tuning of three Lagrange multipliers.

\textbf{Hard constraints: }An alternative approach is to force \e{g(\cdot)} to select one chunck of text with a pre-specified length \e{l}. Instead of predicting the mask directly, \e{g(\cdot)} produces a score \e{\bm s_n} for each position \e{n}, and predicts the start position of the chunk by choosing the maximum of the score. Formally
\begin{equation}
    \small
    n^* = \argmax_{n} \bm s_n, \quad \bm m_n = \mathbbm{1}[n\in [n^*, n^* + l-1]],
    \label{eq:hard_constraint}
\end{equation}
where \e{\mathbbm{1}} denotes the indicator function, which equals \e{1} if the argument is true, and \e{0} otherwise.  Equation~\eqref{eq:hard_constraint} is not differentiable, so when computing the gradients for the back propagation, we apply the straight-through technique \cite{bengio2013estimating} and approximate it with the gradient of
\begin{equation*}
    \small
    \begin{aligned}
    \hat{\bm s} = \textrm{softmax}(\bm s), \quad \bm m = \textrm{CausalConv}(\hat{\bm s}),
    \end{aligned}
\end{equation*}
where \e{\textrm{CausalConv}(\cdot)} denotes causal convolution, and the convolution kernel is an all-one vector of length \e{l}. 

\section{Experiments}
\subsection{Datasets}
To evaluate the invariant rationale generation, we consider the following two binary classification datasets with known spurious correlations.  

\vspace*{-0.05in}
\paragraph{IMDB \cite{maas2011learning}:} The original dataset consists of 25,000 movie reviews for training and 25,000 for testing.  The output \e{Y} is the binarized score of the movie.  We construct a synthetic setting that manually injects tokens with false correlations with \e{Y}, whose prior varies across artificial environments. The goal is to validate if the proposed method \emph{excludes} these tokens from rationale selections.   Specifically, we first randomly split the training set into two balanced subsets, where each subset is considered as an environment.  We append punctuation ``,'' and ``.'' at the beginning of each sentence with the following distributions:
\begin{equation*}
\small
\begin{aligned}
& p(\text{append , } | Y=1, e_i) = p(\text{append . } | Y=0, e_i) = \alpha_i  \\
& p(\text{append . } | Y=1, e_i) = p(\text{append , } | Y=0, e_i) = 1 - \alpha_i.
\end{aligned}
\label{eq:pollution}
\end{equation*}
Here $i$ is the environment index taking values on $\{0, 1\}$. Specifically, we set $\alpha_0$ and $\alpha_1$ to be 0.9 and 0.7, respectively, for the training set.   For the purpose of model selection and evaluation, we randomly split the original test set into two balanced subsets, which are our new validation and test sets.  To test how different rationalization techniques generalize to unknown environments, we also inject the punctuation to the test and validation set, but with $\alpha_0$ and $\alpha_1$ set as 0.5 for the validation set, and 0.1, 0.3 for the testing set.  According to equation~\eqref{eq:pollution}, these manually injected ``,'' and ``.'' can be thought of as the \e{\bm X_2} variable in the figure \ref{fig:model}, which have strong correlations to the label.  It is worth mentioning that the environment ID is only provided in the training set.  

\vspace*{-0.05in}
\paragraph{Multi-aspect beer reviews \cite{mcauley2012learning}:} This dataset is commonly used in the field of rationalization \cite{lei2016rationalizing, bao2018deriving, yu2019rethinking, chang2019game}.  It contains 1.5 million beer reviews, each of which evaluates multiple aspects of a beer.  These aspects include appearance, aroma, smell, palate and overall.  Each aspect has a rating at the scale of [0, 1].  The goal is to provide rationales for these ratings.  There is a high correlation among the rating scores of different aspects in the same review, making it difficult to directly learn a rationalization model from the original data. Therefore only the decorrelated subsets are selected as training data in the previous usages~\cite{lei2016rationalizing,yu2019rethinking}. 

However, the high correlation among rating scores in the original data provides us a perfect evaluation benchmark for \algnamens on its ability to exclude irrelevant but highly correlated aspects, because these highly correlated aspects can be thought of as \e{\bm X_2} and \e{\bm X_3} in figure~\ref{fig:model}, as discussed in section~\ref{subsec:mmi_limit}.  To construct different environments, we cluster the data based on different degree of correlation among the aspects.  To gauge the correlation among aspect, we train a simple linear regression model to predict the rating of the target aspect given the ratings of all the other aspects except the overall.  A low prediction error of the data implies high correlation among the aspects. We then assign the data into different environments based on the linear prediction error. In particular, we construction two training environments using the data with least prediction error, \emph{i.e.} highest correlations. The first training environment is sampled from the lowest 25 percentile of the prediction error, while the second one is from 25 to 50 percentile.  On the contrary, we construction a validation set and a subjective evaluation set from data with the highest prediction error (\emph{i.e.} highest 50 percentile). Following the same evaluation protocol \cite{bao2018deriving, chang2019game}, we consider a classification setting by treating reviews with ratings $\le$ 0.4 as negative and $\ge$ 0.6 as positive.  Each training environment has a total 5,000 label-balanced examples, which makes the size of the training set as 10,000.  The size of the validation set and the subjective evaluation set are 2,000 and 400, respectively.  Same as almost all previous work in rationalization, we focus on the appearance, aroma, and palate aspects only.  

Also, this dataset includes sentence-level annotations for about 1,000 reviews.  Each sentence is annotated with one or multiple aspects label, indicating which aspect this sentence belonging to.  We use this set to automatically evaluate the precision of the extracted rationales.

\subsection{Baselines}

We consider the following two baselines:

{\bf \textsc{Rnp}:} A generator-predictor framework proposed by \citet{lei2016rationalizing} for rationalizing neural prediction (\textsc{Rnp}).  The generator selects text spans as rationales which are then fed to the predictor for label classification. The selection optimizes the MMI criterion shown in equation (\ref{eq:tao}).

{\bf \textsc{3Player}:} The improvement of \textsc{Rnp} from \citet{yu2019rethinking}, which aims to alleviate the degeneration problem of \textsc{Rnp}.  The model consists of three modules, which are the generator, the predictor and the complement predictor.  The complement predictor tries to maximize the predictive accuracy from unselected words.  Besides the MMI objective optimized between the generator and predictor, the generator also plays an adversarial game with the complement predictor, trying to minimize its performance.

\subsection{Implementation Details}
For all experiments, we use bidirectional gated recurrent units \cite{chung2014empirical} with hidden dimension 256 for the generator and both of the predictors.  
All the methods use fixed 100-dimension Glove embeddings~\cite{pennington2014glove}.
We use the Adam optimizer \cite{kingma2014adam} with a learning rate of 0.001.  The batch size is set to 500.  To seek fair comparisons, we try to keep the settings of both \textsc{Rnp} and \textsc{3Player} the same to ours.  We re-implement the \textsc{Rnp}, and use the open-source implementation\footnote{\scriptsize{\url{https://github.com/Gorov/three\_player\_for\_emnlp}.}} of \textsc{3Player}.   The only major difference between these models is that both \textsc{Rnp} and \algname use the straight-through technique \cite{bengio2013estimating} to deal with the problem of nondifferentiability in rationale selections while \textsc{3Player} is based on the policy gradient \cite{williams1992simple}.  

For the IMDB dataset, we follow a standard setting \cite{lei2016rationalizing, chang2019game} to use the soft constraints to regularize the selected rationales for all methods.  Hyperparameters (\emph{i.e.}, $\mu_1$, $\mu_2$ in equation (\ref{eq:soft_constraint}), $\lambda$ in equation (\ref{eq:obj_gen}), and the number of training epochs) are determined based on the best performance on the validation set.   For the beer review task, we find the baseline methods perform much worse using soft constraints compared to the hard one.   This might be because the review of each aspect is highly correlated in the training set.  Thus, we consider the hard constraints with different length in generating rationales.

\subsection{Results}

\paragraph{IMDB: } Table~\ref{tab:imdb} shows the results of the synthetic IMDB dataset.  As we can see, \textsc{Rnp} selects the injected punctuation in 78.24\% of the testing samples, while \algnamens, as expected, does not highlight any.  This result verifies our theoretical analysis in section~\ref{subsec:air}.  Moreover, because \textsc{Rnp} relies on these injected punctuation, whose probabilistic distribution varies drastically between training set and test set, its generalizability is poor,  which leads to low predictive accuracy on the testing set.   Specifically, there is a large gap of around 15\% between the test performance of \textsc{Rnp} and the proposed \algnamens.  It is worth pointing out that, by the dataset construction, \textsc{3Player} will obviously fail by including all punctuation as rationales.  This is because otherwise, the complement predictor will have a clear clue to guess the predicted label.  Thus, we exclude \textsc{3Player} from the comparison.  

\begin{table}[t!]
\small
\centering
\caption{\small{Results on the synthetic IMDB dataset. The last column is the percentage of testing examples with the injected punctuation selected as a part of the rationales. The best test results are {\bf bolded}.}}
\label{tab:imdb}
\begin{tabular}{lccc}
\hline
             & Dev Acc & Test Acc       & Bias Highlighted \\ \hline \hline
\textsc{Rnp} & 78.90   & 72.25          & 78.24           \\
\algname     &  86.65   & {\bf 87.05}    & {\bf 0.00}   \\ \hline
\end{tabular}
\end{table}


\begin{table*}[t!]
\small
\centering
\caption{\small{Experimental results on the multi-aspect beer reviews.  We compare with the baselines on highlight lengths of 10, 20 and 30.  For each aspect and length, we report the best accuracy on the validation set and its corresponding performance on the human annotation set.  The best precision (P), recall (R) and F1 score are {\bf bolded}. }}
\label{tab:beer}
\begin{tabular}{lccccccccccccc}
\hline
\multirow{2}{*}{Methods} & \multirow{2}{*}{Len} & \multicolumn{4}{c}{Appearance}                             & \multicolumn{4}{c}{Aroma}                                  & \multicolumn{4}{c}{Palate}                                 \\
                         &                      & Dev Acc & P              & R              & F1             & Dev Acc & P              & R              & F1             & Dev Acc & P              & R              & F1             \\ \hline \hline
\textsc{Rnp}             & 10                   & 75.20   & 13.51          & 5.75           & 8.07           & 75.30   & 30.30          & 15.26          & 20.30          & 75.00   & 28.20          & 17.24          & 21.40          \\
\textsc{3Player}         & 10                   & 77.55   & 15.84          & 6.78           & 9.50           & 80.75   & \textbf{48.85} & \textbf{24.43} & \textbf{32.57} & 76.60   & 14.15          & 8.54           & 10.65          \\
\algname                 & 10                   & 75.65   & \textbf{49.54} & \textbf{20.93} & \textbf{29.43} & 77.95   & 48.21          & 24.36          & 32.36          & 76.10   & \textbf{32.80} & \textbf{20.01} & \textbf{24.86} \\ \hline
\textsc{Rnp}             & 20                   & 77.70   & 13.54          & 11.29          & 12.31          & 78.85   & 34.32          & 34.18          & 34.25          & 77.10   & 19.80          & 23.78          & 21.60          \\
\textsc{3Player}         & 20                   & 82.56   & 15.63          & 13.47          & 14.47          & 82.95   & 35.73          & 35.89          & 35.81          & 79.75   & 20.73          & 24.91          & 22.63          \\
\algname                 & 20                   & 81.30   & \textbf{58.03} & \textbf{49.59} & \textbf{53.48} & 81.90   & \textbf{42.72} & \textbf{42.52} & \textbf{42.62} & 80.45   & \textbf{44.04} & \textbf{52.75} & \textbf{48.00} \\ \hline
\textsc{Rnp}             & 30                   & 81.65   & 26.26          & 33.10          & 29.29          & 83.10   & 39.97          & 60.13          & 48.02          & 78.55   & 19.18          & 33.81          & 24.47          \\
\textsc{3Player}         & 30                   & 80.55   & 12.56          & 15.90          & 14.03          & 84.40   & 33.02          & 49.66          & 39.67          & 81.85   & 21.98          & 39.27          & 28.18          \\
\algname                 & 30                   & 82.85   & \textbf{54.03} & \textbf{69.23} & \textbf{60.70} & 84.40   & \textbf{44.72} & \textbf{67.35} & \textbf{53.75} & 81.00   & \textbf{26.51} & \textbf{46.91} & \textbf{33.87} \\ \hline
\end{tabular}
\end{table*}

\floatsetup[table]{capposition=bottom}
\begin{table*}[t!]
	\small
	\begin{tabular}{p{\linewidth}}
        \emph{Beer - Appearance} \hspace*{0pt}\hfill Rationale Length - 20\\
		\arrayrulecolor{grey}  
		\midrule
        \hlg{\ul{into}}{0} \hlg{\ul{\textbf{a}}}{20} \hlg{\ul{\textbf{pint}}}{20} \hlg{\ul{\textbf{glass}}}{20} \hlg{\ul{\textbf{,}}}{20} \hlg{\ul{\textbf{poured}}}{20} \hlg{\ul{\textbf{a}}}{20} \hlg{\ul{\textbf{solid}}}{20} \hlg{\ul{\textbf{black}}}{20} \hlg{\ul{\textbf{,}}}{20} \hlg{\ul{\textbf{not}}}{20} \hlg{\ul{\textbf{so}}}{20} \hlg{\ul{\textbf{much}}}{20} \hlg{\ul{\textbf{head}}}{20} \hlg{\ul{\textbf{but}}}{20} \hlg{\ul{\textbf{enough}}}{20} \hlg{\ul{\textbf{,}}}{20} \hlg{\ul{\textbf{tannish}}}{20} \hlg{\ul{\textbf{in}}}{20} \hlg{\ul{\textbf{color}}}{20} \hlg{\ul{\textbf{,}}}{20} \hlg{\ul{decent}}{0} \hlg{\ul{lacing}}{0} \hlg{\ul{down}}{0} \hlg{\ul{the}}{0} \hlg{\ul{glass}}{0} \hlg{\ul{.}}{0} \hlg{as}{0} \hlg{for}{0} \hlg{aroma}{0} \hlg{,}{0} \hlg{if}{0} \hlg{you}{0} \hlg{love}{0} \hlg{coffee}{0} \hlg{and}{0} \hlg{beer}{0} \hlg{,}{0} \hlg{its}{0} \hlg{the}{0} \hlg{best}{0} \hlg{of}{0} \hlg{both}{0} \hlg{worlds}{0} \hlg{,}{0} \hlg{a}{0} \hlg{very}{0} \hlg{fresh}{0} \hlg{strong}{0} \hlg{full}{0} \hlg{roast}{0} \hlg{coffee}{0} \hlg{blended}{0} \hlg{with}{0} \hlg{(}{0} \hlg{and}{0} \hlg{almost}{0} \hlg{overtaking}{0} \hlg{)}{0} \hlg{a}{0} \hlg{solid}{0} \hlg{,}{0} \hlg{classic}{0} \hlg{stout}{0} \hlg{nose}{0} \hlg{,}{0} \hlg{with}{0} \hlg{the}{0} \hlg{toasty}{0} \hlg{,}{0} \hlg{chocolate}{0} \hlg{malts}{0} \hlg{.}{0} \hlg{with}{0} \hlg{the}{0} \hlg{taste}{0} \hlg{,}{0} \hlg{its}{0} \hlg{even}{0} \hlg{more}{0} \hlg{coffee}{0} \hlg{,}{0} \hlg{and}{0} \hlg{while}{0} \hlg{its}{0} \hlg{my}{0} \hlg{dream}{0} \hlg{come}{0} \hlg{true}{0} \hlg{,}{0} \hlg{so}{0} \hlg{delicious}{0} \hlg{,}{0} \hlg{what}{0} \hlg{with}{0} \hlg{its}{0} \hlg{nice}{0} \hlg{chocolate}{0} \hlg{and}{0} \hlg{burnt}{0} \hlg{malt}{0} \hlg{tones}{0} \hlg{again}{0} \hlg{,}{0} \hlg{but}{0} \hlg{i}{0} \hlg{almost}{0} \hlg{say}{0} \hlg{it}{0} \hlg{$<$unknown$>$}{0} \hlg{any}{0} \hlg{$<$unknown$>$}{0} \hlg{,}{0} \hlg{and}{0} \hlg{takes}{0} \hlg{away}{0} \hlg{from}{0} \hlg{the}{0} \hlg{beeriness}{0} \hlg{of}{0} \hlg{this}{0} \hlg{beer}{0} \hlg{.}{0} \hlg{which}{0} \hlg{is}{0} \hlg{n't}{0} \hlg{to}{0} \hlg{say}{0} \hlg{it}{0} \hlg{is}{0} \hlg{n't}{0} \hlg{delicious}{0} \hlg{,}{0} \hlg{because}{0} \hlg{it}{0} \hlg{is}{0} \hlg{,}{0} \hlg{just}{0} \hlg{seems}{0} \hlg{a}{0} \hlg{bit}{0} \hlg{unbalanced}{0} \hlg{.}{0} \hlg{oh}{0} \hlg{well}{0} \hlg{!}{0} \hlg{the}{0} \hlg{mouth}{0} \hlg{is}{0} \hlg{pretty}{0} \hlg{solid}{0} \hlg{,}{0} \hlg{a}{0} \hlg{bit}{0} \hlg{light}{0} \hlg{but}{0} \hlg{not}{0} \hlg{all}{0} \hlg{that}{0} \hlg{unexpected}{0} \hlg{with}{0} \hlg{a}{0} \hlg{coffee}{0} \hlg{blend}{0} \hlg{.}{0} \hlg{its}{0} \hlg{fairly}{0} \hlg{smooth}{0} \hlg{,}{0} \hlg{not}{0} \hlg{quite}{0} \hlg{creamy}{0} \hlg{,}{0} \hlg{well}{0} \hlg{carbonated}{0} \hlg{,}{0} \hlg{thoroughly}{0} \hlg{,}{0} \hlg{exceptionally}{0} \hlg{drinkable}{0} \hlg{.}{0} \\
		\vspace{0.2mm}
		\emph{Beer - Aroma} \hspace*{0pt}\hfill Rationale Length - 20\\
		\arrayrulecolor{grey}  
		\midrule
        \hlr{into}{0} \hlr{a}{0} \hlr{pint}{0} \hlr{glass}{0} \hlr{,}{0} \hlr{poured}{0} \hlr{a}{0} \hlr{solid}{0} \hlr{black}{0} \hlr{,}{0} \hlr{not}{0} \hlr{so}{0} \hlr{much}{0} \hlr{head}{0} \hlr{but}{0} \hlr{enough}{0} \hlr{,}{0} \hlr{tannish}{0} \hlr{in}{0} \hlr{color}{0} \hlr{,}{0} \hlr{decent}{0} \hlr{lacing}{0} \hlr{down}{0} \hlr{the}{0} \hlr{\textbf{glass}}{20} \hlr{\textbf{.}}{20} \hlr{\ul{\textbf{as}}}{20} \hlr{\ul{\textbf{for}}}{20} \hlr{\ul{\textbf{aroma}}}{20} \hlr{\ul{\textbf{,}}}{20} \hlr{\ul{\textbf{if}}}{20} \hlr{\ul{\textbf{you}}}{20} \hlr{\ul{\textbf{love}}}{20} \hlr{\ul{\textbf{coffee}}}{20} \hlr{\ul{\textbf{and}}}{20} \hlr{\ul{\textbf{beer}}}{20} \hlr{\ul{\textbf{,}}}{20} \hlr{\ul{\textbf{its}}}{20} \hlr{\ul{\textbf{the}}}{20} \hlr{\ul{\textbf{best}}}{20} \hlr{\ul{\textbf{of}}}{20} \hlr{\ul{\textbf{both}}}{20} \hlr{\ul{\textbf{worlds}}}{20} \hlr{\ul{\textbf{,}}}{20} \hlr{\ul{a}}{0} \hlr{\ul{very}}{0} \hlr{\ul{fresh}}{0} \hlr{\ul{strong}}{0} \hlr{\ul{full}}{0} \hlr{\ul{roast}}{0} \hlr{\ul{coffee}}{0} \hlr{\ul{blended}}{0} \hlr{\ul{with}}{0} \hlr{\ul{(}}{0} \hlr{\ul{and}}{0} \hlr{\ul{almost}}{0} \hlr{\ul{overtaking}}{0} \hlr{\ul{)}}{0} \hlr{\ul{a}}{0} \hlr{\ul{solid}}{0} \hlr{\ul{,}}{0} \hlr{\ul{classic}}{0} \hlr{\ul{stout}}{0} \hlr{\ul{nose}}{0} \hlr{\ul{,}}{0} \hlr{\ul{with}}{0} \hlr{\ul{the}}{0} \hlr{\ul{toasty}}{0} \hlr{\ul{,}}{0} \hlr{\ul{chocolate}}{0} \hlr{\ul{malts}}{0} \hlr{\ul{.}}{0} \hlr{with}{0} \hlr{the}{0} \hlr{taste}{0} \hlr{,}{0} \hlr{its}{0} \hlr{even}{0} \hlr{more}{0} \hlr{coffee}{0} \hlr{,}{0} \hlr{and}{0} \hlr{while}{0} \hlr{its}{0} \hlr{my}{0} \hlr{dream}{0} \hlr{come}{0} \hlr{true}{0} \hlr{,}{0} \hlr{so}{0} \hlr{delicious}{0} \hlr{,}{0} \hlr{what}{0} \hlr{with}{0} \hlr{its}{0} \hlr{nice}{0} \hlr{chocolate}{0} \hlr{and}{0} \hlr{burnt}{0} \hlr{malt}{0} \hlr{tones}{0} \hlr{again}{0} \hlr{,}{0} \hlr{but}{0} \hlr{i}{0} \hlr{almost}{0} \hlr{say}{0} \hlr{it}{0} \hlr{$<$unknown$>$}{0} \hlr{any}{0} \hlr{$<$unknown$>$}{0} \hlr{,}{0} \hlr{and}{0} \hlr{takes}{0} \hlr{away}{0} \hlr{from}{0} \hlr{the}{0} \hlr{beeriness}{0} \hlr{of}{0} \hlr{this}{0} \hlr{beer}{0} \hlr{.}{0} \hlr{which}{0} \hlr{is}{0} \hlr{n't}{0} \hlr{to}{0} \hlr{say}{0} \hlr{it}{0} \hlr{is}{0} \hlr{n't}{0} \hlr{delicious}{0} \hlr{,}{0} \hlr{because}{0} \hlr{it}{0} \hlr{is}{0} \hlr{,}{0} \hlr{just}{0} \hlr{seems}{0} \hlr{a}{0} \hlr{bit}{0} \hlr{unbalanced}{0} \hlr{.}{0} \hlr{oh}{0} \hlr{well}{0} \hlr{!}{0} \hlr{the}{0} \hlr{mouth}{0} \hlr{is}{0} \hlr{pretty}{0} \hlr{solid}{0} \hlr{,}{0} \hlr{a}{0} \hlr{bit}{0} \hlr{light}{0} \hlr{but}{0} \hlr{not}{0} \hlr{all}{0} \hlr{that}{0} \hlr{unexpected}{0} \hlr{with}{0} \hlr{a}{0} \hlr{coffee}{0} \hlr{blend}{0} \hlr{.}{0} \hlr{its}{0} \hlr{fairly}{0} \hlr{smooth}{0} \hlr{,}{0} \hlr{not}{0} \hlr{quite}{0} \hlr{creamy}{0} \hlr{,}{0} \hlr{well}{0} \hlr{carbonated}{0} \hlr{,}{0} \hlr{thoroughly}{0} \hlr{,}{0} \hlr{exceptionally}{0} \hlr{drinkable}{0} \hlr{.}{0} \\
		\vspace{0.2mm}
		\emph{Beer - Palate} \hspace*{0pt}\hfill Rationale Length - 20\\
		\arrayrulecolor{grey}  
		\midrule
        \hlb{into}{0} \hlb{a}{0} \hlb{pint}{0} \hlb{glass}{0} \hlb{,}{0} \hlb{poured}{0} \hlb{a}{0} \hlb{solid}{0} \hlb{black}{0} \hlb{,}{0} \hlb{not}{0} \hlb{so}{0} \hlb{much}{0} \hlb{head}{0} \hlb{but}{0} \hlb{enough}{0} \hlb{,}{0} \hlb{tannish}{0} \hlb{in}{0} \hlb{color}{0} \hlb{,}{0} \hlb{decent}{0} \hlb{lacing}{0} \hlb{down}{0} \hlb{the}{0} \hlb{glass}{0} \hlb{.}{0} \hlb{as}{0} \hlb{for}{0} \hlb{aroma}{0} \hlb{,}{0} \hlb{if}{0} \hlb{you}{0} \hlb{love}{0} \hlb{coffee}{0} \hlb{and}{0} \hlb{beer}{0} \hlb{,}{0} \hlb{its}{0} \hlb{the}{0} \hlb{best}{0} \hlb{of}{0} \hlb{both}{0} \hlb{worlds}{0} \hlb{,}{0} \hlb{a}{0} \hlb{very}{0} \hlb{fresh}{0} \hlb{strong}{0} \hlb{full}{0} \hlb{roast}{0} \hlb{coffee}{0} \hlb{blended}{0} \hlb{with}{0} \hlb{(}{0} \hlb{and}{0} \hlb{almost}{0} \hlb{overtaking}{0} \hlb{)}{0} \hlb{a}{0} \hlb{solid}{0} \hlb{,}{0} \hlb{classic}{0} \hlb{stout}{0} \hlb{nose}{0} \hlb{,}{0} \hlb{with}{0} \hlb{the}{0} \hlb{toasty}{0} \hlb{,}{0} \hlb{chocolate}{0} \hlb{malts}{0} \hlb{.}{0} \hlb{with}{0} \hlb{the}{0} \hlb{taste}{0} \hlb{,}{0} \hlb{its}{0} \hlb{even}{0} \hlb{more}{0} \hlb{coffee}{0} \hlb{,}{0} \hlb{and}{0} \hlb{while}{0} \hlb{its}{0} \hlb{my}{0} \hlb{dream}{0} \hlb{come}{0} \hlb{true}{0} \hlb{,}{0} \hlb{so}{0} \hlb{delicious}{0} \hlb{,}{0} \hlb{what}{0} \hlb{with}{0} \hlb{its}{0} \hlb{nice}{0} \hlb{chocolate}{0} \hlb{and}{0} \hlb{burnt}{0} \hlb{malt}{0} \hlb{tones}{0} \hlb{again}{0} \hlb{,}{0} \hlb{but}{0} \hlb{i}{0} \hlb{almost}{0} \hlb{say}{0} \hlb{it}{0} \hlb{$<$unknown$>$}{0} \hlb{any}{0} \hlb{$<$unknown$>$}{0} \hlb{,}{0} \hlb{and}{0} \hlb{takes}{0} \hlb{away}{0} \hlb{from}{0} \hlb{the}{0} \hlb{beeriness}{0} \hlb{of}{0} \hlb{this}{0} \hlb{beer}{0} \hlb{.}{0} \hlb{which}{0} \hlb{is}{0} \hlb{n't}{0} \hlb{to}{0} \hlb{say}{0} \hlb{it}{0} \hlb{is}{0} \hlb{n't}{0} \hlb{delicious}{0} \hlb{,}{0} \hlb{because}{0} \hlb{it}{0} \hlb{is}{0} \hlb{,}{0} \hlb{just}{0} \hlb{seems}{0} \hlb{a}{0} \hlb{bit}{0} \hlb{unbalanced}{0} \hlb{.}{0} \hlb{oh}{0} \hlb{well}{0} \hlb{!}{0} \hlb{\ul{the}}{0} \hlb{\ul{\textbf{mouth}}}{20} \hlb{\ul{\textbf{is}}}{20} \hlb{\ul{\textbf{pretty}}}{20} \hlb{\ul{\textbf{solid}}}{20} \hlb{\ul{\textbf{,}}}{20} \hlb{\ul{\textbf{a}}}{20} \hlb{\ul{\textbf{bit}}}{20} \hlb{\ul{\textbf{light}}}{20} \hlb{\ul{\textbf{but}}}{20} \hlb{\ul{\textbf{not}}}{20} \hlb{\ul{\textbf{all}}}{20} \hlb{\ul{\textbf{that}}}{20} \hlb{\ul{\textbf{unexpected}}}{20} \hlb{\ul{\textbf{with}}}{20} \hlb{\ul{\textbf{a}}}{20} \hlb{\ul{\textbf{coffee}}}{20} \hlb{\ul{\textbf{blend}}}{20} \hlb{\ul{\textbf{.}}}{20} \hlb{\ul{\textbf{its}}}{20} \hlb{\ul{\textbf{fairly}}}{20} \hlb{\ul{smooth}}{0} \hlb{\ul{,}}{0} \hlb{\ul{not}}{0} \hlb{\ul{quite}}{0} \hlb{\ul{creamy}}{0} \hlb{\ul{,}}{0} \hlb{\ul{well}}{0} \hlb{\ul{carbonated}}{0} \hlb{\ul{,}}{0} \hlb{\ul{thoroughly}}{0} \hlb{\ul{,}}{0} \hlb{\ul{exceptionally}}{0} \hlb{\ul{drinkable}}{0} \hlb{\ul{.}}{0} \\
	\end{tabular}
	\vspace*{-0.15in}
    \captionof{figure}{\small{Examples of \algname generated rationales on the multi-aspect datasets.  Human annotated words are \ul{underlined}. Appearance, aroma and palate rationales are in bold text and highlighted in \hlg{\textbf{green}}{20}, \hlr{\textbf{red}}{20}, and \hlb{\textbf{blue}}{20} respectively. }}
    \label{fig:exp_highlight}
\end{table*}
\floatsetup[table]{capposition=top}

\vspace*{-0.05in}
\paragraph{Beer review: }
We conduct both objective and subjective evaluations for the beer review dataset.  We first compare the generated rationales against the human annotations and report precision, recall and F1 score in table \ref{tab:beer}.   Similarly, the reported performances are based on the best performance on the validation set, which is also reported.  We consider the highlight lengths of 10, 20 and 30. 

We observe that \algname consistently surpass the other two baselines in finding rationales that align with human annotation for most of the rationale lengths and the aspects.   In particular, although the best accuracies among all three methods on validation sets have only small variations, the improvements are significant in terms of finding the correct rationales.  For example, \algname improves over the other two methods for more than 20 absolute percent in F1 for the appearance aspect.  Two baselines methods fail to distinguish the true clues for different aspects, which confirms that the previous MMI objective is insufficient for ruling out the spurious words.  

\begin{figure}[t]
\centering
\includegraphics[width=\linewidth]{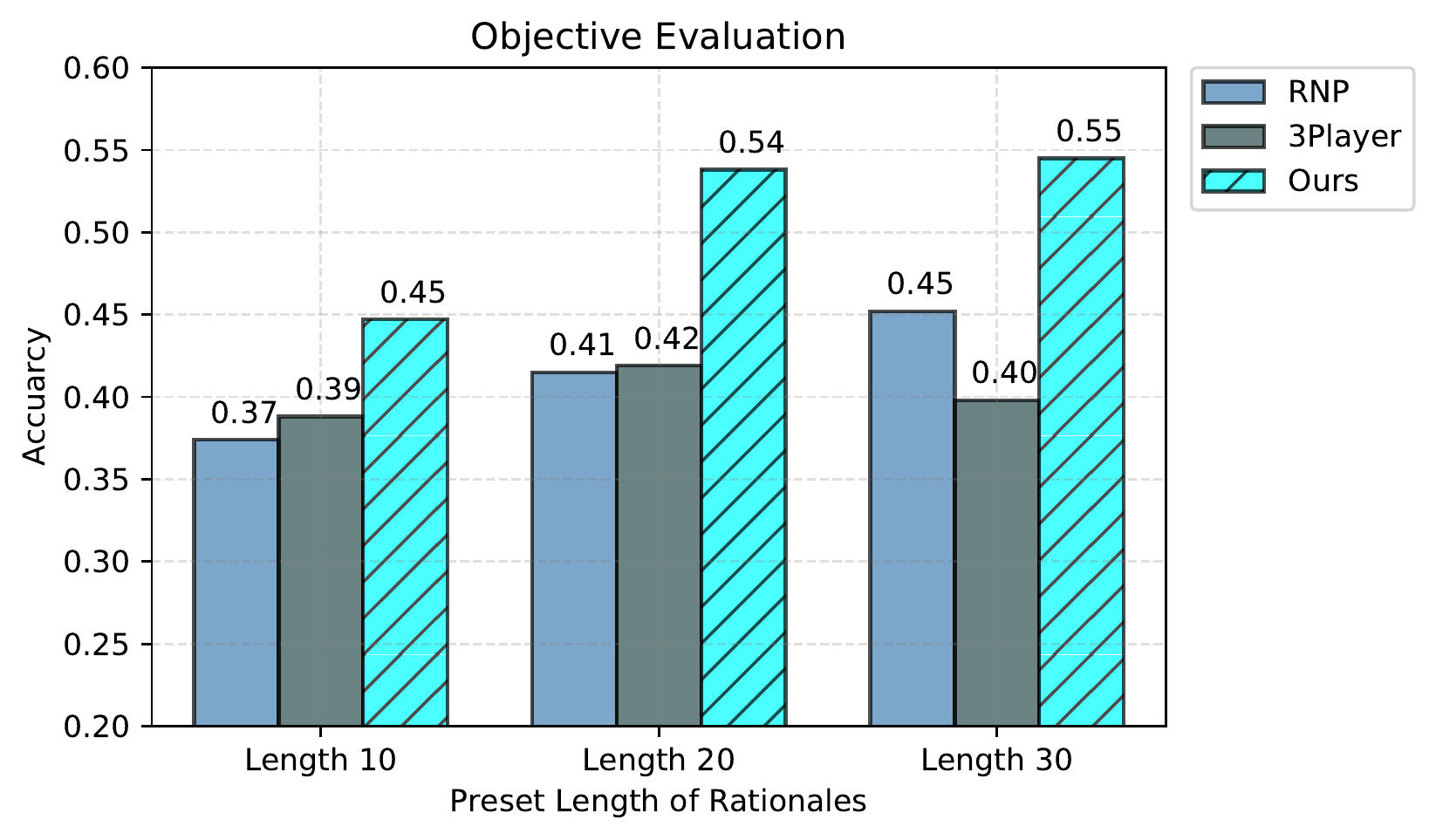}
\vspace*{-0.35in}
\caption{\small{Subjective performances of generated rationales.  Subjects are asked to guess the target aspect (\emph{i.e.} which aspect of the model is trained on) based on the generated rationales.  We report the case of preset rationale length of 10, 20 and 30. }}
\label{fig:human_evaluation}
\end{figure}

In addition, we also visualize the generated rationales of our method with a preset length of 20 in figure \ref{fig:exp_highlight}.  We observe that the \algname is able to produce meaningful justifications for all three aspects.  By reading these selected texts alone, humans will easily predict the aspect label.  To further verify that the rationales generated by \algname align with human judgment, we present a subjective evaluation via \emph{Amazon Mechanical Turk}.  Recall that for each aspect we preserved a hold-out set with 400 examples (total 1,200 examples for all three aspects).  We generate rationales with different lengths for all methods. In each subjective test, the subject is presented with the rationale of one aspect of the beer review, generated by one of the three methods (unselected words blocked), and asked to guess which aspect the rationale is talking about.  We then compute the accuracy as the performance metric, which is shown in figure \ref{fig:human_evaluation}.  Under this setting, a generator that picks spurious correlated texts will have a low accuracy.   As can be observed, \algname achieves the best performances in all cases.

\section{Related Work}

\paragraph{Selective rationalization:}  Selective rationalization is one of the major categories of model interpretability in machine learning.  \citet{lei2016rationalizing} first propose a generator-predictor framework for rationalization.  The framework is formally a co-operative game that maximizes the mutual information between the selected rationales and labels, as shown in~\cite{chen2018learning}.   Following this work, \citet{chen2018shapley} improves the generator-predictor framework by proposing a new rationalization criterion by considering the combinatorial nature of the selection.  \citet{yu2019rethinking} point out the communication problem in co-operative learning and proposes a new three-player framework to control the unselected texts.  \citet{chang2019game} aim to generate rationales in all possible classes instead of the target label only, which makes the model perform counterfactual reasoning.  In all, these models deal with different challenges in generating high-quality rationales.  However, they are still insufficient to distinguish the invariant words from the correlated ones.  

\paragraph{Self-explaining models beyond selective rationalization:}
Besides selective rationalization, other approaches also improve the interpretability of neural predictions.  For example,  module networks \cite{andreas2016learning,andreas2016neural,johnson2017inferring} compose appropriate modules following the logical program produced by a natural language component.  The restriction to a small set of pre-defined programs currently limits their applicability.  Other lines of work include evaluating feature importance with gradient information \cite{simonyan2013deep,li2016visualizing, sundararajan2017axiomatic} or local perturbations \cite{kononenko2010efficient, lundberg2017unified};  and interpreting deep networks by locally fitting interpretable models~\cite{ribeiro2016should,alvarez2018towards}.  However, these methods aim at providing post-hoc explanations of already-trained models, which is not able to find invariant texts.  


\paragraph{Learning with biases:}
Our work also relates to the topic of discovering dataset-specific biases.  Specifically, neural models have shown remarkable results in many NLP applications, however, these models sometimes prone to fit some dataset-specific patterns or biases.  For example, in natural language inference, such biased clues can be the word overlap between the input sentence pair \cite{mccoy2019right} or whether the negative word "\emph{not}" exists \cite{niven2019probing}.  Similar observations have been found in multi-hop question answering \cite{welbl2018constructing, min2019compositional}.  To learn with biased data but not fully rely on it,  \citet{lewis2018generative} use generative objectives to force the QA models to make use of the full question.  \citet{agrawal2018don,wang2019multi} propose carefully designed model architectures to capture more complex interactions between input clues beyond the biases.  \citet{ramakrishnan2018overcoming,belinkov2019adversarial} propose to add adversarial regularizations that punish the internal representations that cooperate well with bias-only models. \citet{clark2019don,he2019unlearn} propose to learn ensemble models that fit the residual from the prediction with bias features.  However, all these works assume that the biases are known.  Our work instead can rule out unwanted features without knowing the pattern priorly.  
\section{Conclusion}
In this paper, we propose a game-theoretic approach to invariant rationalization, where the method is trained to constrain the probability of the output conditional on the rationales be the same across multiple environments.  The framework consists of three players, which competitively rule out spurious words with strong correlations to the output.  We theoretically demonstrate the proposed game-theoretic framework drives the solution towards better generalization to test scenarios that have different distributions from the training.  Extensive objective and subjective evaluations on both synthetic and multi-aspect sentiment classification datasets demonstrate that \algname performs favorably against existing algorithms in rationale generation.

%
{
\bibliographystyle{icml2019}
\bibliography{main}
}
\appendix
\section{Proof To Theorem \ref{thm:invariance}}
\label{appendix:proof}
\begin{proof}
\e{\forall \bm Z}, partition \e{\bm Z} into an invariant variable \e{\bm Z_I} and a non-invariant variable \e{\bm Z_V}:
\begin{equation*}
    \small
    \begin{aligned}
    \bm Z_I = \bm Z \cap \{ \bm X_1 \}, \quad \bm Z_V = \bm Z \cap \{ \bm X_2, \bm X_3 \}.
    \end{aligned}
\end{equation*}

Given an arbitrary \e{\pi_1}, we construct a specific \e{\pi_2=\pi_2^*} and \e{\pi_3=\pi_3^*} such that
\begin{equation}
    \small
    \begin{aligned}
    &\pi_2^*(\bm x_2 | \bm x_1, y) = \pi_2^*(\bm x_2), \quad \pi_3^*(\bm x_3 | \bm x_1) = \pi_3^*(\bm x_3).
    \end{aligned}
    \label{eq:proof_assump}
\end{equation}
In other words, set these two priors such that the all the non-invariant variables are uninformative of \e{Y}. Since the test adversary is allowed to choose any distribution, this set of priors is within the feasible set of the test adversary.

Under the set of priors in equation (\ref{eq:proof_assump}), the non-invariant features are not predicative of \e{Y}, and only the invariant features are predicative of \e{Y}, \emph{i.e.}
\begin{equation}
    \small
    p(Y | \bm Z, e_a) = p(Y | \bm Z_I, e_a)
    \label{eq:proof1}
\end{equation}
Therefore
\begin{equation}
    \small
    \begin{aligned}
    \mathcal{L}^*_\textrm{test}(\bm Z; \pi_1, \pi_2^*, \pi_3^*) &= H(p(Y | \bm Z, e_a); p(Y | \bm Z, e_t)) \\
    &\overset{(i)}{=} H(p(Y | \bm Z_I, e_a); p(Y | \bm Z, e_t)) \\
    &\overset{(ii)}{\geq}  H(p(Y | \bm Z_I, e_a)) \\
    &\overset{(iii)}{\geq}  H(p(Y | \bm X_1, e_a)) \\
    &\overset{(iv)}{=}  H(p(Y | \bm X_1, e_a); p(Y | \bm X_1, e_t)) \\
    &= \mathcal{L}^*_\textrm{test}(\bm X_1; \pi_1, \pi_2^*, \pi_3^*)
    \end{aligned}
    \label{eq:proof2}
\end{equation}
where (i) is from equation \eqref{eq:proof1}; (ii) is from the relationship between cross entropy and entropy; (iii) is because \e{\bm X_1} is the minimizer of conditional entropy of \e{Y} on \e{\bm Z_I} and \e{e_a}, among all the invariant variables; (iv) is because, by the definition of invariant variables, \e{p(Y | \bm X_1, E) = p(Y | \bm X_1)}.  Here, we use \e{\mathcal{L}^*_\textrm{test}(\bm Z; \pi_1, \pi_2^*, \pi_3^*)} to emphasize that \e{\mathcal{L}^*_\textrm{test}(\bm Z)} is computed under the distribution of \e{\pi_1, \pi_2^*, \pi_3^*}.  Therefore, if we optimize over \e{\pi_2} and \e{\pi_3}, we have the following
\begin{equation}
\small
\begin{aligned}
    &\max_{\pi_2, \pi_3} \mathcal{L}^*_\textrm{test}(\bm Z; \pi_1, \pi_2, \pi_3) \geq \mathcal{L}^*_\textrm{test}(\bm Z; \pi_1, \pi_2^*, \pi_3^*),\\
    &\max_{\pi_2, \pi_3} \mathcal{L}^*_\textrm{test}(\bm X_1; \pi_1, \pi_2, \pi_3) = \mathcal{L}^*_\textrm{test}(\bm X_1; \pi_1, \pi_2^*, \pi_3^*)
\end{aligned}
\label{eq:proof3}
\end{equation}
where the second line is because \e{p(Y|\bm X_1, e_a)} does not depend on \e{\pi_2} and \e{\pi_3}.  Combining equations \eqref{eq:proof2} and \eqref{eq:proof3}, we have
\begin{equation}
\small
\begin{aligned}
    \max_{\pi_2, \pi_3} \mathcal{L}^*_\textrm{test}(\bm Z; \pi_1, \pi_2, \pi_3) \geq \max_{\pi_2, \pi_3} \mathcal{L}^*_\textrm{test}(\bm X_1; \pi_1, \pi_2, \pi_3)
\end{aligned}
\label{eq:proof4}
\end{equation}
Note that the above discussions holds for all \e{\pi_1}. Therefore, taking the maximum over \e{\pi_1} of equation \eqref{eq:proof4} preserves the inequality.
\begin{equation*}
\small
\begin{aligned}
    \max_{\pi_1, \pi_2, \pi_3} \mathcal{L}^*_\textrm{test}(\bm Z; \pi_1, \pi_2, \pi_3) \geq \max_{\pi_1, \pi_2, \pi_3} \mathcal{L}^*_\textrm{test}(\bm X_1; \pi_1, \pi_2, \pi_3)
\end{aligned}
\label{eq:proof5}
\end{equation*}
which implies
\begin{equation*}
    \small
    \begin{aligned}
    \bm X_1 = \argmin_{\bm Z} \max_{\pi_1, \pi_2, \pi_3} \mathcal{L}^*_\textrm{test}(\bm Z; \pi_1, \pi_2, \pi_3)
    \end{aligned}
\end{equation*}
\end{proof}

\end{document}